\documentclass[10pt,journal,compsoc]{IEEEtran}

\ifCLASSOPTIONcompsoc
  \usepackage[nocompress]{cite}
\else
  \usepackage{cite}
\fi
\ifCLASSINFOpdf
  \usepackage[pdftex]{graphicx}
\else
\fi
\usepackage{amsmath,amsfonts}
\usepackage{bm}

\ifCLASSOPTIONcompsoc
 \usepackage[caption=false,font=footnotesize,labelfont=sf,textfont=sf]{subfig}
\else
 \usepackage[caption=false,font=footnotesize]{subfig}
\fi

\usepackage{color}
\usepackage[table]{xcolor}
\usepackage{multirow}
\usepackage{tikz}
\usepackage{booktabs}
\usepackage{pgfplots}
\usepackage{textcomp}
\usepackage{algorithm}
\usepackage{algorithmic}
\usepackage{pifont}
\usepackage{verbatim}
\usepackage{graphicx}
\usepackage{stfloats}

\usepackage{booktabs}
\usepackage{multirow}
\usepackage{makecell}
\usepackage{tikz}
\usepackage{color, colortbl}

\pgfplotsset{compat=1.18}
\makeatletter
\newcommand{\thickhline}{%
	\noalign {\ifnum 0=`}\fi \hrule height 1pt
	\futurelet \reserved@a \@xhline
}
\makeatother

\definecolor{mygray}{gray}{.9}

\begin{document}
%
\title{MV2DFusion: Leveraging Modality-Specific Object Semantics for Multi-Modal 3D Detection}
%
%
%
%

\author{Zitian~Wang, Zehao~Huang, Yulu~Gao, Naiyan~Wang, Si~Liu$^\dag$
\IEEEcompsocitemizethanks{
\IEEEcompsocthanksitem $\dag$ Corresponding author. 
\IEEEcompsocthanksitem Zitian Wang, Yulu Gao and Si Liu are with Institute of Artificial Intelligence, Beihang University (e-mail:
{\texttt{wangzt.kghl@gmail.com, gyl97@buaa.edu.cn, liusi@buaa.edu.cn}}).
\IEEEcompsocthanksitem Zehao Huang and Naiyan Wang (e-mail: 
\texttt{zehaohuang18@gmail.com},\texttt{winsty@gmail.com}).

}}

%
%

\markboth{IEEE TRANSACTIONS ON PATTERN ANALYSIS AND MACHINE INTELLIGENCE}{}
\IEEEtitleabstractindextext{%
\begin{abstract}
The rise of autonomous vehicles has significantly increased the demand for robust 3D object detection systems. While cameras and LiDAR sensors each offer unique advantages—cameras provide rich texture information and LiDAR offers precise 3D spatial data—relying on a single modality often leads to performance limitations. This paper introduces MV2DFusion, a multi-modal detection framework that integrates the strengths of both worlds through an advanced query-based fusion mechanism. By introducing an image query generator to align with image-specific attributes and a point cloud query generator, MV2DFusion effectively combines modality-specific object semantics without biasing toward one single modality. Then the sparse fusion process can be accomplished based on the valuable object semantics, ensuring efficient and accurate object detection across various scenarios. Our framework's flexibility allows it to integrate with any image and point cloud-based detectors, showcasing its adaptability and potential for future advancements. Extensive evaluations on the nuScenes and Argoverse2 datasets demonstrate that MV2DFusion achieves state-of-the-art performance, particularly excelling in long-range detection scenarios. 
\end{abstract}

}

\maketitle

\IEEEdisplaynontitleabstractindextext

%
\IEEEpeerreviewmaketitle

\ifCLASSOPTIONcompsoc
\IEEEraisesectionheading{\section{Introduction}\label{sec:introduction}}
\else
\section{Introduction}
\label{sec:introduction}
\fi

\IEEEPARstart{T}{he} rise of autonomous vehicles has heightened the demand for 3D object detection.
The differing imaging principles of sensors, such as cameras and LiDAR, allow the corresponding data modalities to capture different attributes of real-world objects.
The inherent characteristics of different modalities make them excel at distinguishing objects from different perspectives.
For instance, in images the objects are presented as texture-rich pixel regions, while in point clouds the objects are presented as groups of 3D points. 
In recent years, many efforts have been made in camera-based detection~\cite{BEVDet,BEVFormer,DETR3D,PETR} and LiDAR-based detection~\cite{pvrcnn,second,pointpillars,voxelnet} and achieve great progress.
Nevertheless, detection relying on a single modality exposes its own deficiency.
For instance, images lack depth information to indicate the 3D positions, while point clouds lack rich semantic information and struggle to capture distant objects due to sparsity.

To leverage the strengths of both modalities, multi-modal fusion approaches have been proposed, promising to provide the best of both worlds.
Current multi-modal fusion methods can be broadly categorized into two main strategies: feature-level fusion and proposal-level fusion.
Feature-level fusion methods construct a unified feature space (typically in LiDAR frame), where different modality features are extracted to form a multi-modal feature volume. DeepFusion~\cite{deepfusion} and AutoAlign~\cite{autoalign,autoalignv2} use point cloud features to query image features to augment the point cloud feature representation. BEVFusion~\cite{bevfusion_mit,bevfusion_pku} transform both image and point cloud features into BEV space and fuse them together.
CMT~\cite{yan2023cross} does not construct a unified feature space but adopts a unified feature aggregation approach of attention~\cite{vaswani2017attention} to query both image and point cloud features.
Though feature-level fusion methods make it straightforward to accomplish object identification and localization, they do not fully exploit the object priors embedded in the original modality data. Some may even impair the strong modality-specific semantic information during the fusion process~\cite{fu2024eliminating}.

Differently, proposal-level fusion methods utilize modality-specific proposals to make the most of modality data.
One of the pioneering methods F-PointNet~\cite{f-pointnet} transforms detected image bounding boxes into frustums to retrieve objects from point clouds. FSF~\cite{FSF} and SparseFusion~\cite{sparsefusion} first produce image proposals and point cloud proposals from each modality, then unify them into point cloud based instance representations for multi-modal interaction.
However, in both methods, the representations are biased toward one modality. For instance, camera proposals dominate the multi-modal fusion process in the former, while in the latter, the image proposals are essentially transformed into the same representation of point cloud proposals. 

To address the above challenges, we present a multi-modal detection framework named MV2DFusion in this paper.
Our framework extends MV2D~\cite{mv2d} to incorporate multi-modal detection, whose object-as-query design facilitates a natural extension to multi-modal settings. 
We re-engineer the image query generator to be more aligned with the image modality, where uncertainty-aware image queries are introduced to preserve the object semantics from the images and inherit the rich semantics as in the projection view.
By introducing an additional point cloud query generator, we can also obtain the object semantics from point clouds and combine them with image queries. Then the fusion process can be conducted easily in an attention-based format.

With the deliberate query generator design, we can fully leverage the modality-specific object semantics without being tied to one specific representation space.
Moreover, it also allows us to integrate any type of image detector and point cloud detector, showcasing the framework's versatility and capacity for advancement. 
Thanks to the sparse nature of the fusion strategy, our framework is also feasible to deploy in long-range scenarios without quadratically increased memory consumption and computational cost.
Additionally, with minimal modifications, this framework can easily incorporate any query-based methods to utilize historical information effectively (e.g., StreamPETR~\cite{streampetr}).

We evaluate our proposed method on large-scale 3D detection benchmarks nuScenes~\cite{nuscenes2019} and Argoverse 2~\cite{argoverse2} (AV2), achieving state-of-the-art performances. Our contributions can be summarized as follows:
\begin{itemize}
    \item We propose a framework that fully leverages modality-specific object semantics for comprehensive multi-modal detection. The effectiveness and efficiency of the framework are validated in the nuScenes and AV2 datasets.
    \item The framework can be flexibly implemented with any modality detectors depending on the deployment environment and can harvest the continuously evolving detection models for better performance.
    \item The framework provides a feasible solution in the long-range scenarios thanks to the sparsity of the fusion strategy. 
\end{itemize}

In conclusion, our approach demonstrates advancements in multi-modal 3D detection, offering a robust and versatile solution that leverages the strengths of both camera and LiDAR modalities.

\section{Related Work}

\subsection{LiDAR-based 3D Detection}
LiDAR-based 3D detection has drawn much attention in the field of autonomous driving because it can provide accurate depth and structure information. Current methods are mainly divided into point-based~\cite{shi2019pointrcnn,yang20203dssd,FSD,fsdv2}, voxel-based~\cite{yan2018second,yin2021center,transfusion}, pillar-based~\cite{pointpillars}, and range-based~\cite{fan2021rangedet,sun2021rsn}.
Point-based methods directly utilize the original point cloud without quantization. PointRCNN~\cite{shi2019pointrcnn} employs a two-stage network, where stage one creates 3D proposals directly from the raw point cloud and stage two refines the proposals by fusing semantic features and local spatial features. 
3DSSD~\cite{yang20203dssd} builds a lightweight and efficient point-based 3D single-stage object detection framework by adopting a novel sampling strategy based on 3D Euclidean distance and feature distance. 
FSD~\cite{FSD} introduces a point-based module to address the issue of missing center features in a fully sparse pipeline.
The voxel-based method quantizes the original point into voxels. Due to the sparsity of voxel space, SECOND~\cite{yan2018second} introduces sparse convolution in processing voxel representation to reduce training and inference cost by a large margin. Based on the voxel space, CenterPoint~\cite{yin2021center} proposes a powerful center-based anchor-free 3D detector, which becomes a widely used baseline method for 3D object detection.
Transfusion-L~\cite{transfusion} introduces Transformer~\cite{vaswani2017attention} into the detection head to improve higher performance.
Range-based~\cite{fan2021rangedet,sun2021rsn} and pillar-based~\cite{pointpillars} methods  convert the original point cloud to a 2D representation from different perspective and use the 2D network for feature extraction. 

In this paper, the LiDAR branch of our proposed framework can be applied to mainstream LiDAR detectors such as FSD~\cite{FSD}, VoxelNeXt~\cite{voxelnext}, and TransFusion-L~\cite{transfusion}. When utilizing fully sparse detectors, our entire framework retains its fully sparse characteristics.

\subsection{Camera-based 3D Detection}
Due to the lower cost of the camera sensors, camera based 3D detections also attracted great attentions.
2D detectors have been developed for many years, and a direct way is to lift 2D detectors into 3D detectors. Following this way, CenterNet~\cite{zhou2019objects} and FCOS3D~\cite{FCOS3D} convert 3D targets into the image domain for supervision, so that the 3D box can be directly predicted in the perspective view.
Some methods explicitly build a BEV space for prediction, and they project image features onto the BEV space with depth information from depth estimation~\cite{wang2019pseudo} or directly query 2d image features by using 3D-2D cross-attention technology.
BEVDet~\cite{BEVDet}, BEVDet4D~\cite{huang2022bevdet4d} and BEVDepth~\cite{BEVDepth} utilize the LSS~\cite{LSS} module to project features from multiple camera views into a BEV  representation by leveraging the predicted depth distribution.
BEVFormer~\cite{BEVFormer} uses spatial cross-attention to perform 2D-to-3D transformations and also utilizes temporal self-attention to represent the current environment by incorporating historical BEV features.
PolarFormer~\cite{jiang2022polarformer} advocates the exploitation of the Polar coordinate system and proposes a new Polar Transformer for more accurate 3D object detection in the BEV.
While these 3D space representations are beneficial for unifying multi-view images, the memory consumption and computational cost increase as the detection range in 3D space expands.

Some methods follow DETR~\cite{DETR} by using queries to aggregate image features for prediction. DETR3D~\cite{DETR3D} utilizes a sparse set of 3D object queries to index into these 2D features, without estimating dense 3D scene geometry. PETR~\cite{PETR,petrv2}  directly assigns 3D position embeddings to 2D images. 
Sparse4D~\cite{lin2022sparse4dv1,lin2023sparse4dv2,lin2023sparse4dv3} leverages 4D reference points to sample features across multiple frames without relying on dense view transformation.
It is worth noting that the aforementioned methods adopt learnable object queries, requiring dense object queries distributed in 3D space to ensure sufficient recall of objects. MV2D~\cite{mv2d} proposes a 2D object-guided framework that leverages dynamic queries to recall objects and eliminate interference from noise and distractors. In this paper, we extend MV2D~\cite{mv2d} to incorporate multi-modal detection, and our framework maintains the feasibility of being deployed in long-range scenarios avoiding quadratic memory consumption and computational cost.
Additionally, our framework can easily incorporate any query-based temporal modeling methods like StreamPETR \cite{streampetr} to effectively utilize historical information.

\subsection{Fusion-based 3D Detection}

Cameras and LiDAR are complementary to each other. The camera contains dense color and texture information while LiDAR provides precise depth and structure information, so they can cooperate with each other for higher precision. Therefore, multi-modal fusion becomes a key issue to promote the performance and robustness of the perceptual system. Current multi-modal fusion methods are mainly divided into feature-level fusion and proposal-level fusion. Feature-level fusion methods aim to construct a unified feature space. Early approaches~\cite{vora2020pointpainting,wang2021pointaugmenting,yin2021multimodal,xu2021fusionpainting} project the point cloud onto the image and append semantic labels around the projection location to the 3D points. More recent methods, such as SD-Fusion~\cite{gao2023sparse} and AutoAlign~\cite{autoalign,autoalignv2}, use point cloud features to query image features, thereby enhancing the representation of point cloud features. BEVFusion~\cite{bevfusion_mit,bevfusion_pku} transforms both image and point cloud features into the BEV space and fuses them. 
CMT~\cite{yan2023cross} integrates 3D coordinates into image and point cloud tokens, allowing the DETR~\cite{DETR} pipeline to be used effectively for multi-modal fusion and end-to-end learning.
SparseFusion~\cite{sparsefusion_tusimple}, on the other hand, focuses on sparsifying BEV features to improve the efficiency of the fusion process. Proposal-level fusion methods exploit modality-specific proposals to maximize the utility of modality data. F-PointNet~\cite{f-pointnet}, converts detected image bounding boxes into frustums to extract objects from point clouds. FSF~\cite{FSF} and SparseFusion~\cite{sparsefusion} first generate proposals from images and point clouds separately, then unify them into point cloud-based instance representations for multi-modal interaction. However, these methods often bias toward one modality, which makes the methods cannot fully harvest the benefit of fusion.
In this paper, we propose a proposal-level approach leveraging modality-specific object semantics to address the issue of modality bias.

\section{Methodology} 
\begin{figure*}[t!]
	\centering
	\vspace{-4.5mm}
	\includegraphics[width=1.0\textwidth]{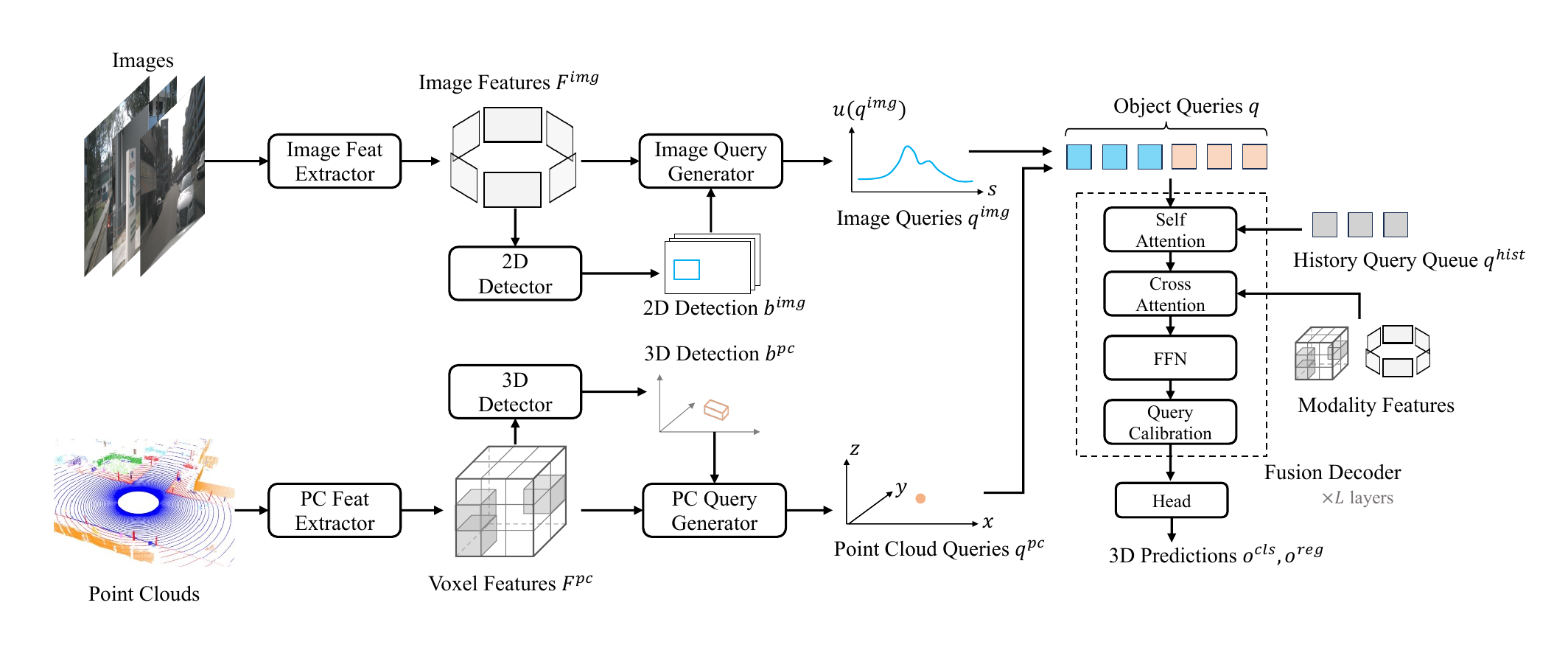}
	\caption{The framework of the proposed MV2DFusion. The model processes multi-view images and point clouds as inputs, extracting modality features via an image backbone and a point cloud backbone independently. Simultaneously, an image-based 2D detector and a point cloud-based 3D detector are applied to these features, yielding respective detection results. Subsequently, image queries and point cloud queries are generated by respective query generators based on the modality features and detection results. Finally, these queries and features are fed into a fusion decoder, where the queries integrate information from both modalities, leading to 3D predictions after query updating.
}
	\vspace{-3.5mm}
	\label{fig:framework}
\end{figure*}

\subsection{Overview}

The overall pipeline of MV2DFusion is shown in Figure~\ref{fig:framework}.
Given $N^{img}$ multi-view images and point clouds as inputs, the model first extracts modality specific features through an image backbone and a point cloud backbone independently.
An image-based 2D detector and a point cloud-based 3D detector are applied to the modality features and output respective detection results. 
Then image queries and point cloud queries are generated by corresponding query generators based on the modality features and detection results.
Finally, the modality queries and features are fed into a fusion decoder, where the queries aggregate information from both modalities, and make 3D predictions after query updating.
We will elaborate the details and design principles of each part in the following sections.

\subsection{Exploiting Modality-Specific Object Semantics}
We devise a fusion strategy that unearths the information in the original forms of modalities without biasing one modality to another.
Particularly, rather than representing and fusing the complete scene in 3D space, we tackle multi-modal 3D detection by exploiting and fusing the modality-specific object semantics.
This fusion strategy preserves the unique characteristics of each modality, while also providing the benefit of sparsity to reduce computational costs and memory usage.

\subsubsection{Object Proposals from Modality-Specific Experts}
\label{sec:feature_extraction}
First, we adopt independent backbones for modality feature extraction.
The image backbone with FPN~\cite{FasterRCNN} neck extracts image features $\{\mathbf{F}^{img}_v | 1 \leq v \leq N^{img}\}$ from the multi-view images. The LiDAR backbone extracts voxel features $\mathbf{F}^{pc}$ from the point clouds. During feature extraction, the two branches do not interact with each other to maintain the distinct information in each modality.

Based on the modality features, we leverage the \emph{modality-specific experts} to discover the proposals of objects within each modality. Each expert may explore the characteristics of each modality for better performance.
For instance, image-based 2D detections mainly rely on rich texture on pixels, while point cloud-based 3D detection focus on the point distribution induced shapes of the objects.

Specifically, for image-based 2D detection, we can use any 2D detector with limiting the structure of the detector, e.g., anchor-based~\cite{RetinaNet} or anchor-free~\cite{yolox2021}, two-stage~\cite{FasterRCNN} or one-stage~\cite{FCOS}. The 2D detection produces $M^{img}$ 2D bounding boxes for every image, with each box represented by $(x_{min}, y_{min}, x_{max}, y_{max})$. 
The overall 2D detection results can be described as $\{\mathbf{b}^{img}_v \in \mathbb{R}^{M^{img}\times 4}|1\leq v \leq N^{img}\}$.

Thanks to the proposed sparse fusion strategy, for point cloud-based 3D detection, we can use sparse detectors~\cite{voxelnext,FSD,fsdv2} directly operating on voxels to make the whole model \emph{fully sparse}~\cite{FSD}.\footnote{Our model is also compatible with detectors that accept BEV features.}
Since fully sparse models do not construct dense BEV features, they have significant advantages in reducing memory usage and computational cost, especially under long-range scenarios~\cite{FSD}. 
The 3D detections consist of $M^{pc}$ 3D bounding boxes, with each box represented as $(x, y, z, w, l, h, rot)$. The overall 3D detection results can be described as $\mathbf{b}^{pc} \in \mathbb{R}^{M^{pc}\times 7}$.

\subsubsection{Deriving Object-level Semantics from Experts}
Though both the detection results provide valuable cues for identifying objects, their representations are essentially different. The point cloud-based 3D detections are presented in the 3D space, while the image-based 2D detections are presented in the projected 2D space.
This large domain gap causes difficulty in fusing the information together.  
In this paper, we propose to derive object-level semantics from the detection, rather than directly fusing the raw detection results.

As point clouds are typically distributed along the surfaces of objects, they excel at accurately capturing the objects' shape and pose.
But unlike in point clouds, an object's 3D pose cannot be directly inferred from the images.
Instead, how the objects are distributed in the image plane can serve as a clue for 3D localization considering the projection principles.  
On the other hand, the image pixels can describe objects with rich textures, even over long distances where the sparse point clouds might fail to capture objects. 

Considering the different characteristics, we adopt the form of \emph{object query} to encode the object-level semantics from each modality, then the multi-modal information can be seamlessly integrated.
In modern transformer-based detection frameworks~\cite{anchor-detr,dn-detr,dino}, each object query typically consists of two components: the content part and the positional part. 
After obtaining the detection results from both the image-based detection results 
$\{\mathbf{b}^{img}_v\}$ and the point cloud-based detection results $\mathbf{b}^{pc}$, we construct object queries based on the bounding boxes and corresponding modality features. 
We will detail the object query generation process for each modality in the following sections.

\begin{figure}[t!]
	\centering
	\includegraphics[width=0.5\textwidth]{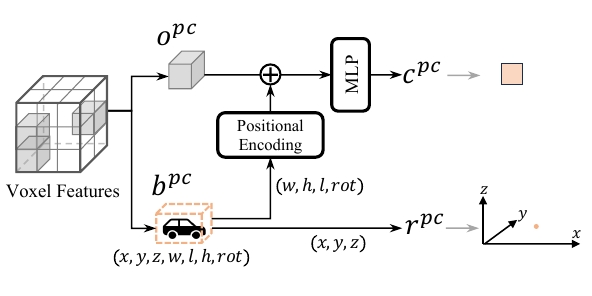}
	\vspace{-6mm}
	\caption{Illustration of point cloud query generator. Some superscripts and subscripts are omitted for simplicity.}
	\label{fig:pc_query_generator}
\end{figure}

\subsubsection{Point Cloud Object Query Generation}
\label{sec:pc_query_generation}
Since the point cloud-based 3D detector directly outputs in the 3D space, we adopt the object center points $\mathbf{r}^{pc}\in \mathbb{R}^{M^{pc}\times 3}$ that can indicate the factual locations of presented objects as positional parts. For the content parts $\mathbf{c}^{pc}\in \mathbb{R}^{M^{pc}\times C}$, we integrate both appearance and geometric features. 
Therefore, the point cloud queries are represented as:
\begin{equation}
\label{equ:point_cloud_query}
    \mathbf{q}^{pc} = (\mathbf{c}^{pc}, \mathbf{r}^{pc}).
\end{equation}

The details of point cloud query generator are shown in Figure~\ref{fig:pc_query_generator}.
It is noteworthy that the appearance features $\mathbf{o}^{pc}$ depend on the kind of detector. For instance, in a center-based detector~\cite{second}, the appearance features are the values in each BEV grid. In a two-stage detector~\cite{pvrcnn}, the appearance features are the RoI features.
In our implementation, we use a sparse point cloud-based 3D detector~\cite{fsdv2}. So the appearance features $\mathbf{o}^{pc}$ are the voxel features from which the predictions are drawn. 
We consider the geometric features as objects' physical properties like size and heading, which are explicitly represented in detection results $\mathbf{b}^{pc}$.
So the content parts $\mathbf{c}^{pc}$ can be briefly formulated as:
\begin{equation}
    \mathbf{c}^{pc} = \text{MLP}(\mathbf{o}^{pc} + \text{MLP}(\text{SinPos}(\mathbf{b}^{pc}))),
\end{equation}
where $\text{SinPos}$ denotes the sinusoidal positional encoding~\cite{vaswani2017attention} to transform low-dimensional vectors into high-dimensional features.

\begin{figure}[t!]
	\centering
	\includegraphics[width=0.5\textwidth]{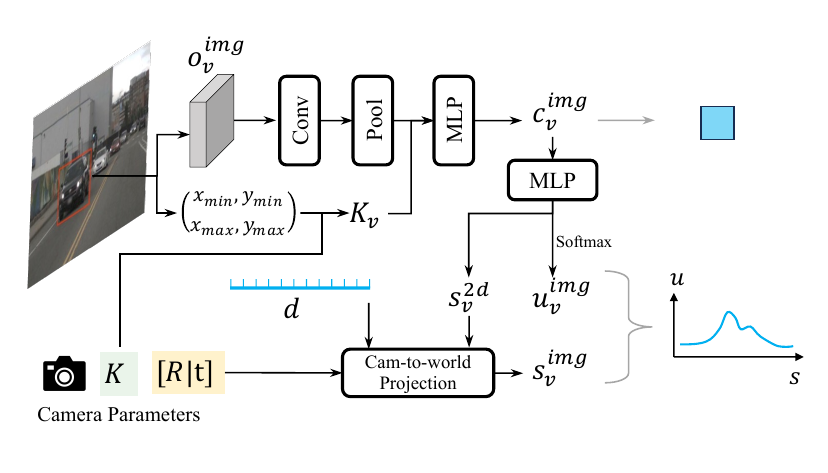}
	\vspace{-6mm}
	\caption{Illustration of image query generator. Some superscripts and subscripts are omitted for simplicity.}
	\label{fig:img_query_generator}
\end{figure}

\subsubsection{Image Object Query Generation}
\label{sec:img_query_generation}
An intuitive way to generate image queries is to apply an image-based 3D detector and transform the detected instances into image queries~\cite{sparsefusion}. 
By this way, one can readily obtain image queries in the same formation as point cloud queries, facilitating multi-model fusion between queries.
However, due to the inherent differences between modalities, forcing the same form for different modality queries would hurt the performance. 
Due to the ill-posed problem of depth estimation from images, the 3D predictions are susceptible to large errors, which will hamper both the feature quality and localization quality of image queries.

To handle this challenge, we introduce novel uncertainty-aware image queries based on 2D detections.
For the queries derived from the $v$-th camera view, the content parts $\mathbf{c}^{img}_v \in \mathbb{R}^{M^{img}\times C}$ are the RoI (region of interest) appearance features embedded with geometric information. As for positional parts, we do not opt for the estimated object centers like previous works~\cite{mv2d} but retain the uncertainty of depth estimation from images.
Concretely, we formulate the query positions by probability distributions rather than deterministic values. The distributions are modeled using categorical distributions, consisting of $n_d$ sampling positions $\mathbf{s}^{img} \in \mathbb{R}^{M^{img}\times n_{d}\times3}$ and their corresponding probabilities $\mathbf{u}^{img}\in \mathbb{R}^{M^{img}\times n_d}$.

With the content parts $\mathbf{c}^{img}_v$ and positional parts $\mathbf{s}^{img}, \mathbf{u}^{img}$, image queries from the $v$-th camera view are represented as $\mathbf{q}^{img}_v=(\mathbf{c}^{img}_v, \mathbf{s}^{img}_v,\mathbf{u}^{img}_v)$.
The overall image queries $\mathbf{q}^{img}$ are aggregated from each camera view:
\begin{equation}
\label{equ:image_query}
    \mathbf{q}^{img} = \{\mathbf{q}^{img}_v=(\mathbf{c}^{img}_v, \mathbf{s}^{img}_v,\mathbf{u}^{img}_v)|1\leq v \leq N^{img}\}.
\end{equation}

Given this formulation of uncertainty-aware image queries, we can provide an initial guess of the object locations~\cite{mv2d,bevformerv2} and alleviate the 
inaccuracies brought by camera-to-world projection. The detailed process is illustrated in Figure~\ref{fig:img_query_generator}.

Given the 2D object detection results $\mathbf{b}^{img}_v$ and image feature maps $\mathbf{F}^{img}_v $ from the $v$-th image, the image query generator first extracts the RoI object appearance features $\mathbf{o}_v^{img} \in \mathbb{R}^{M_v^{img} \times H^{r} \times W^{r} \times C}$ through RoI-Align~\cite{MaskRCNN}, where $H^{r}\times W^{r}$ denotes the spatial size:
\begin{equation}
    \mathbf{o}_v^{img} = \text{RoI-Align}(\mathbf{F}^{img}_v, \mathbf{b}^{img}_v).
\end{equation}

Besides the appearance features $\mathbf{o}_v^{img}$, we also feed equivalent camera intrinsic matrices $\mathbf{K}_v$ into the image query generator to compensate for the geometric information lost during RoI-Align~\cite{mv2d}. Denote the original camera intrinsic matrix is:
\begin{equation}
\mathbf{K}_v^{ori} =
{
\begin{bmatrix}
    f_x  & 0 & o_x &  0\\
    0 & f_y & o_y & 0\\
    0 & 0 & 1 & 0 \\
    0 & 0 & 0 & 1
\end{bmatrix}},
\end{equation}
then the equivalent camera intrinsic matrix $\mathbf{K}_{v}^i$ that defines the projection from the camera coordinate frame to the $i$-th 2D bounding box $(x_{min}^i, y_{min}^i, x_{max}^i, y_{max}^i)$ can be formulated as:
\begin{equation}
    \mathbf{K}_{v}^i = 
    {
    \begin{bmatrix}
        f_x * r_x & 0 & (o_x-x_{min}^i)*r_x & 0\\
        0 & f_y * r_y & (o_y-y_{min}^i)*r_y & 0\\
        0 & 0 & 1 & 0 \\
        0 & 0 & 0 & 1
    \end{bmatrix}}
    ,
\end{equation}
where $r_x = W^{r} / (x_{max}^i - x_{min}^i), r_y = H^{r} / (y_{max}^i - y_{min}^i)$.

Therefore, the content parts  $\mathbf{c}_{v}^{img}$ of image queries are parameterized by the appearance features $\mathbf{o}_v^{img}$ and the geometry information $\mathbf{K}_{v}$:
\begin{equation}
    \mathbf{c}_{v}^{img} = \text{MLP}(\left[\text{Pool}(\text{Conv}(\mathbf{o}_v^{img}));\text{Flat}(\mathbf{K}_v\right)]),
\end{equation}
where $;$ denotes the concatenating operation and $\text{Flat}()$ means to flatten the trailing dimensions of a tensor.

As for the positional parts, we uniformly sample $n_{d}$ values within a predefined depth range $[d_{min}, d_{max}]$, forming a depth set $\mathbf{d} \in \mathbb{R}^{n_{d}}$. Then we predict a set of 2D sampling positions $\mathbf{s}^{2d} \in \mathbb{R}^{M^{img}\times n_{d}\times2}$ and corresponding probabilities $\mathbf{u}^{img}\in \mathbb{R}^{M^{img}\times n_d}$, where the subscript $v$ is omitted for simplicity:
\begin{align}
    [\mathbf{s}^{2d};\mathbf{u}^{logit}] = \text{MLP}(\mathbf{c}^{img}),\\
    \mathbf{u}^{img} = \text{softmax}(\mathbf{u}^{logit}).
\end{align}
With 2D sampling positions $\mathbf{s}^{2d}$ and depth values $\mathbf{d}$, we can obtain the 3D sampling positions $\mathbf{s}^{img}$ through through camera-to-world projection.

Note that the distribution-based formulation shares some similarities with LSS~\cite{LSS}, but it does not actually distribute the query features into 3D space. 
This encoding format not only saves the computational resources and memories required by LSS, but also improves the robustness when the depth predictions are erroneous.
Moreover, it provides a second chance to refine the locations further afterward, as will be discussed in Section~\ref{sec:query_calibration}. 

\subsection{Fusing Modality Information}
Inspired by the detection transformer~\cite{DETR}, we utilize a decoder-like structure~\cite{vaswani2017attention} to fuse modality information and predict the final results.
The decoder has $L$ decoder layers, composed of self-attention layers, cross-attention layers, layer normalizations, feed-forward networks, and query calibration layers. 
The point cloud queries $\mathbf{q}^{pc}$ and image queries $\mathbf{q}^{img}$ are combined as the input to the decoder.
These combined input queries are noted by $\mathbf{q}^0=(\mathbf{q}^{pc}, \mathbf{q}^{img})$, while the queries after the $l$-th layer are noted by $\mathbf{q}^l$.

The important adaptations are emphasized below.

\subsubsection{Self-Attention}
As mentioned in Equation~\ref{equ:point_cloud_query} and Equation~\ref{equ:image_query}, the modality queries have different formulations, i.e., $\mathbf{q}^{pc}=(\mathbf{c}^{pc}, \mathbf{r}^{pc})$ and $\mathbf{q}^{img}=(\mathbf{c}^{img}, \mathbf{s}^{img},\mathbf{u}^{img})$.
To make them compatible with the typical self-attention layers~\cite{vaswani2017attention}, we keep the content parts and transform the positional parts into a coherent representation.

We utilize the positional encoding method (PE) and the uncertainty-aware positional encoding method (U-PE) to create positional encodings
$\mathbf{p}^{pc}\in \mathbb{R}^{M^{pc}\times C}$ and $\mathbf{p}^{img}\in \mathbb{R}^{M^{img}\times C}$ for each modality respectively:
\begin{align}
\label{equ:PE}
    \mathbf{p}^{pc} &= \text{PE}(\mathbf{r}^{pc}),\\
    \mathbf{p}^{img} &= \text{U-PE}(\mathbf{s}^{img},\mathbf{u}^{img}).
\end{align}

In PE, $\mathbf{p}^{pc} $ are generated from the center points $\mathbf{r}^{pc}$:
\begin{equation}
    \text{PE}(\mathbf{r}^{pc}) = \text{MLP}(\text{SinPos}(\mathbf{r}^{pc})),
\end{equation}
where $\text{SinPos}$ denotes the sinusoidal positional encoding.

In U-PE, we first transform $\mathbf{s}^{img}$ into basic positional encodings $\mathbf{p}^{base}$, then inject the propability $\mathbf{u}^{img}$ into $\mathbf{p}^{img}_{base}$ through a gating operation:
\begin{align}
    \mathbf{s}^{base} &= \text{MLP}(\text{Flat}(\mathbf{s}^{img})),\\
    \text{U-PE}(\mathbf{s}^{img},\mathbf{u}^{img}) &= \text{MLP}(\mathbf{s}^{base} \odot\sigma(\text{MLP}(\mathbf{u}^{img}))),
\end{align}
where $\text{Flat}()$ denotes the flattening operation,
$\odot$ denotes element-wise multiplication and $\sigma$ denotes sigmoid function. 

Given the typical denotation $\text{MHA}(Q, K, V)$ of the multi-head attention, the self-attention can be written as:
\begin{equation}
    \text{SelfAttn} = \text{MHA}(\mathbf{W}^Q(\mathbf{c}^{sa}+\mathbf{p}^{sa}), \mathbf{W}^K(\mathbf{c}^{sa}+\mathbf{p}^{sa}), \mathbf{W}^V\mathbf{c}^{sa}),
\end{equation}
where $\mathbf{p}^{sa}$ stands for the concatenated positional encodings of $\mathbf{p}^{pc}$ and $\mathbf{p}^{img}$, so as $\mathbf{c}^{sa}$.
Without loss of generality, this formulation can be briefly  presented as:
\begin{equation}
\label{equ:self_attn}
    \text{SelfAttn} = \text{MHA}(\mathbf{W}^Q\mathbf{q}^{sa}, \mathbf{W}^K\mathbf{q}^{sa}, \mathbf{W}^V\mathbf{q}^{sa}).
\end{equation}

\begin{figure*}[t!]
	\centering
	\includegraphics[width=1.0\textwidth]{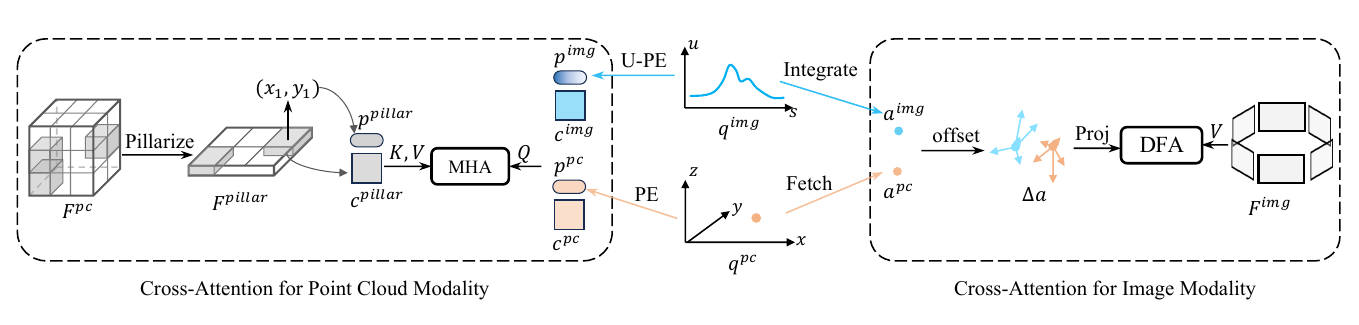}
	\caption{Demonstration of the cross-attention layer. Both point cloud queries and image queries aggregate multi-modal features for useful information. These two kinds of queries undergo respective processing before they interact with features. For more details, please refer to Section.~\ref{sec:catten}. 
}
	\vspace{-4mm}
	\label{fig:cross-attn}
\end{figure*}

\subsubsection{Cross-Attention}
\label{sec:catten}
Though simply exchanging information among queries themselves in self-attention layers yields decent predictions, we found in experiments that adding cross-attention layers as in the transformer decoder can bring further performance gains with the cost of the heavier model. The cross-attention layer targets aggregating useful modality specific features for updating queries as illustrated in Figure~\ref{fig:cross-attn}. 

For image features, we employ the projection-based deformable attention~\cite{Deformable-DETR, BEVFormer}. Specifically, we first obtain the anchor point of each query. The anchor points for point cloud queries $\mathbf{a}^{pc}$  are directly their 3D locations $\mathbf{r}^{pc}$. While the anchor points for image queries $\mathbf{a}^{img}$ are the weighted average of the sampling positions $\mathbf{s}^{img}$ by the probability distributions $\mathbf{u}^{img}$. For the $i$-th image query:
\begin{equation}
    \mathbf{a}^{img}_i = \mathbf{u}^{img\top}_{i} \cdot  \mathbf{s}^{img}_{i}.
\end{equation}

The projection-based deformable attention is calculated by:
\begin{equation}
    \text{DFA}(\mathbf{c}^{m}, \mathbf{a}^{m}, \mathbf{F}^{img}) = 
    \sum_{v=1}^{N^{img}}\sum_{k=1}^K A_{vk} \cdot \mathbf{W} \mathbf{F}^{img}_v(\text{Proj}( \mathbf{a}^{m}+\Delta\mathbf{a}_k)),
\end{equation}
where attention weights $A_{vk}$ and offsets $\Delta\mathbf{a}_k$ are predicted from content parts $\mathbf{c}^{m}$, $K$ denotes the number of sampling points, $\text{Proj}()$ denotes the world-to-camera projection, $m$ denotes the query modality (image or point cloud).

For point cloud features, if the model accepts BEV-formed point cloud features, similar deformable attention operations can be used to exchange information.

In our implementation, the point cloud features $\mathbf{F}^{pc}$ are formed by sparse voxels to make a fully sparse model~\cite{FSD}. In this case, we employ vanilla multi-head attention to aggregate point cloud features. $\mathbf{F}^{pc}$ are compressed by pillarization~\cite{pointpillars}, e.g., average pooling along height, to produce content parts $\mathbf{c}^{pillar}$. The positional encodings $\mathbf{p}^{pillar}$ of pillar features are generated from the BEV positions $\mathbf{r}^{pillar}$:
\begin{equation}
    \mathbf{p}^{pillar} = \text{MLP}(\text{SinPos}(\mathbf{r}^{pillar})).
\end{equation}

\subsubsection{Query Calibration}
\label{sec:query_calibration}
By referring to other queries and modality features, one query can have a more precise description of the object it belongs to. 
Due to the characteristics of different modalities, we consider the query positions generated from images to be less reliable, whereas those generated from point clouds are relatively accurate.
Therefore, we calibrate the image queries with the updated features after each decoder layer. 
The calibration will refine the locations of the image queries and mitigate the uncertainty. 
In the process, only the probability distributions $\mathbf{u}^{img}$ are refined, leaving the sampling positions $\mathbf{s}^{img}$ untouched.

Particularly, the latest $\mathbf{u}^{img}$ are refined from the old one by residual addition. The refinement is accomplished in the logit space rather than the probability space. This process can be written as:
\begin{align}
    \mathbf{u}^{logit} =& \log(\mathbf{u}^{img}),\\
    \mathbf{u}^{img} \xleftarrow{}& \text{softmax}(\mathbf{u}^{logit} + \text{MLP}(\mathbf{c}^{img})).
\end{align}

The query calibration layer will affect the positional encodings $\mathbf{p}^{img}$ and the anchor points $\mathbf{a}^{img}$ accordingly.

\subsubsection{Model Output}
Upon obtaining the object queries $\mathbf{q}^L$ at the last decoder layer, a classification head and a regression head are applied to $\mathbf{q}^L$ to get the model output.
Specifically, the classification scores $\mathbf{z}^{cls}$ can be obtained by the context features $\mathbf{c}^L$:
\begin{equation}
    \mathbf{z}^{cls} = \text{MLP}(\mathbf{c}^L).
\end{equation}

The regression target is the location, size, rotation, and velocity (if needed), formed as $(x, y, z, w, l, h, rot, vx, vy)$.
The regression is made based on the context features $\mathbf{c}^L$ and anchor points $\mathbf{a}^L$:
\begin{equation}
    \mathbf{z}^{reg} = \text{MLP}(\mathbf{c}^L) + [\mathbf{a}^L; \mathbf{0}].
\end{equation}

\subsubsection{Temporal Information Utilization} 
To further harvest the temporal cues, we choose an efficient query-based temporal fusion method with negligible computational cost~\cite{streampetr}.

Specifically, we maintain a history query queue $\mathbf{q}^{hist} \in \mathbb{R}^{(K\times T)\times C}$ of size $(K\times T)$ to serve as history information. 
After finishing the prediction in the current frame, we push the top-$K$ scored object queries into $\mathbf{q}^{hist}$. That makes $\mathbf{q}^{hist}$ contain history information of $T$ previous frames.

The history query queue $\mathbf{q}^{hist}$ is involved in the self-attention layer. Specifically, it is first transformed with corresponding time lags $\mathbf{t}\in \mathbb{R}^{(K\times T)}$, ego poses $\mathbf{P}\in \mathbb{R}^{(K\times T)\times 4\times4}$ and velocities $\mathbf{v}\in \mathbb{R}^{(K\times T)\times 2}$ of history queries:
\begin{equation}
    \mathbf{q}^{trans} = \phi(\mathbf{q}^{hist}|\mathbf{t},\mathbf{P}, \mathbf{v}),
\end{equation}
where $\phi$ is a small network to encode related information. Then the transformed history queries $\mathbf{q}^{trans}$ serve as additional keys and values in the self-attention. The self-attention in Equation~\ref{equ:self_attn} is changed to:
\begin{equation}
    \text{SelfAttn} = \text{MHA}(\mathbf{W}^Q\mathbf{q}^{sa}, \mathbf{W}^K[\mathbf{q}^{sa}; \mathbf{q}^{trans}], \mathbf{W}^V[\mathbf{q}^{sa}; \mathbf{q}^{trans}]).
\end{equation}

\subsection{Loss Functions}
The modality-specific query generators are compatible with arbitrary image-based 2D detectors and LiDAR-based 3D detectors. These detectors can be seamlessly integrated into the MV2DFusion framework without any modifications neither to the model structure nor to the loss function. 
These detectors are pretrained to provide better initialization of dynamic queries. They also can be jointly trained during the training phase. 
In this context, we refer $\mathcal{L}_{det2D}$ to the original loss function of the image-based 2D detector, and $\mathcal{L}_{det3D}$ to that of the LiDAR-based 3D detector.

For the output of fusion decoder layers, we inherit the design of target assignment and loss function from DETR~\cite{DETR}.
Hungarian algorithm~\cite{kuhn1955hungarian} is adopted for label assignment. Focal loss~\cite{RetinaNet} and $L_1$ loss are adopted for object classification and box regression respectively. The output 3D object detection loss can be summarized as:
\begin{equation}
    \mathcal{L}_{out} = \lambda_{cls} \cdot \mathcal{L}_{cls} + \mathcal{L}_{reg}.
\end{equation}

Apart from the above losses for object detection, we add auxiliary supervision on the image query generator to facilitate the depth estimation.
Given $\mathbf{b}_{v}^{img}$ as the 2D bounding boxes predicted from the $v$-th image,  $\mathbf{\hat{b}}_{v}^{proj}$ as the 2D bounding boxes projected from ground-truth 3D bounding boxes, we first calculate the pairwise IoU matrix $U$ between $\mathbf{b}_{v}^{img}$ and  $\mathbf{\hat{b}}_{v}^{proj}$, such that $U_{i,j}=\text{IoU}(\mathbf{b}_{v,i}^{img}, \mathbf{\hat{b}}_{v,j}^{proj})$. The bounding box  $\mathbf{b}_{v,i}^{img}$ is assigned to a target $\mathbf{\hat{b}}_{v,j}^{proj}$ if \texttt{(1)} $U_{ij}=\max\limits_k U_{ik}$, \texttt{(2)} $U_{ij}=\max\limits_k U_{kj}$ and \texttt{(3)} $U_{ij}>\tau_{IoU}$. If the 2D RoI is successfully assigned to the target, the depth distribution output by the image query generator $\mathbf{d}_{v,i}^{img}$ is supervised by the target depth $\mathbf{\hat{d}}_{v,j}^{proj}$ in the auxiliary loss $\mathcal{L}_{aux}$:

\begin{equation}
    \mathcal{L}_{aux}=\text{CELoss}(\mathbf{d}_{v,i}^{img}, \mathbf{\hat{d}}_{v,j}^{proj}),
\end{equation}
where \text{CELoss} refers to cross entropy loss.
The overall loss function of MV2DFusion is:
\begin{equation}
    \mathcal{L} = \lambda_{det2D} \cdot \mathcal{L}_{det2D} + \lambda_{det3d} \cdot \mathcal{L}_{det3d} + \lambda_{aux} \cdot \mathcal{L}_{aux}  + \lambda_{out} \cdot \mathcal{L}_{out}.
\end{equation}
The hyperparameters $\{\lambda\}$ are set empirically to balance the scales of different loss terms.

\section{Experiments}
\subsection{Datasets}
\subsubsection{nuScenes} We conduct experiments on the nuScenes dataset~\cite{nuscenes2019} with 1000 scenes. Each instance within this dataset includes RGB imagery captured by 6 cameras, providing a 360° horizontal field of view, along with  LiDAR. There are 1.4 million 3D bounding boxes across 10 different categories.
We employ the metrics supplied by the dataset to evaluate our results, namely the mean average precision (mAP) and the nuScenes detection score (NDS). 

\subsubsection{Argoverse2} We conduct extensive long-range experiments on the Argoverse 2 (AV2) dataset~\cite{argoverse2} to demonstrate the superiority of our model in long-range detection tasks. AV2 is a large-scale dataset with a perception range of up to 200 meters, covering an area of 400m × 400m. 
It contains 1000 sequences in total, with 700  for training, 150 for validation, and 150 for testing. Each sequence is recorded using seven high-resolution cameras at 20Hz and one LiDAR sensor at 10Hz. For evaluation, in addition to the mean average precision (mAP) metric, AV2 adopts a composite detection score (CDS), which incorporates both AP and localization errors. 

\begin{table*}[t]
\begin{center}
\caption{Performance on the nuScenes \texttt{test} sets. ``-E'' means with model ensemble and TTA. Our model is marked in \colorbox[rgb]{0.8627,0.8627,0.8627}{gray}. \textbf{Bold} indicates the best performance.}
\begin{tabular}{l|cc|ccccc}
\toprule
         {Method}    & NDS $\uparrow$ & mAP $\uparrow$ & \scriptsize{mATE} $\downarrow$  & \scriptsize{mASE} $\downarrow$ & \scriptsize{mAOE} $\downarrow$ & \scriptsize{mAVE}$ \downarrow$ & \scriptsize{mAAE} $\downarrow$   \\
         \midrule

        
         {PointPainting~\cite{vora2020pointpainting}}  & 0.610 & 0.541 &0.380 & 0.260 & 0.541 & 0.293 & 0.131 \\
         {PointAugmenting~\cite{wang2021pointaugmenting}}  & 0.711 & 0.668 & 0.253 &0.235 & 0.354 & 0.266 & 0.123 \\
         {MVP~\cite{yin2021multimodal}}  & 0.705 & 0.664 & 0.263 & 0.238 &0.321 & 0.313 & 0.134 \\
         {TransFusion~\cite{transfusion}}  & 0.717 & 0.689 & 0.259 &0.243 &0.359 &0.288 &0.127 \\
         {UVTR~\cite{li2022unifying}}  &0.711 &0.671 &0.306 &0.245 &0.351 &0.225 &0.124 \\
         {Autoalignv2}~\cite{autoalignv2}  & 0.724 &0.684 & - & - & - & - & - \\
         {BEVFusion}~\cite{bevfusion_mit} & 0.729 & 0.702 & 0.261 & 0.239 & 0.329 & 0.260 & 0.134\\
         {BEVFusion}~\cite{bevfusion_pku}   & 0.733 & 0.713 & 0.250 & 0.240 & 0.359 & 0.254 & 0.132 \\
         {DeepInteraction}~\cite{yang2022deepinteraction}   & 0.734 & 0.708 & 0.257 & 0.240 & 0.325 & 0.245 & 0.128 \\
         {CMT}~\cite{yan2023cross}   &0.741 &0.720 &0.279 &0.235 &0.308 &0.259 &0.112\\
         {EA-LSS}~\cite{hu2023ea}  &0.744 &0.722 &0.247 &0.237 &0.304 &0.250 &0.133  \\
         {BEVfusion4D}~\cite{cai2023bevfusion4d}   & 0.747 & 0.733 & - & - & - & - & -\\
         {SparseFusion}~\cite{zhou2023sparsefusion} &0.738 &0.720 & - & - & - & - & -\\
         {FusionFormer}~\cite{hu2023fusionformer} &0.751 & 0.726 & 0.267 & 0.236 & 0.286 & 0.225 & 0.105\\
         \rowcolor[RGB]{220,220,220} {MV2DFusion}   &\textbf{0.767} &\textbf{0.745} &0.245 &0.229 &0.269 &0.199 &0.115  \\
         
         \hline
         {BEVFusion-E}~\cite{bevfusion_mit}  &0.761 &0.750 &0.242 &0.227 &0.320 &0.222 &0.130 \\
         {DeepInteraction-E}~\cite{yang2022deepinteraction}  &0.763 &0.756 &0.235 &0.233 &0.328 &0.226 &0.130 \\
         {CMT-E}~\cite{yan2023cross}  &0.770 &0.753 &0.233 &0.220 &0.271 &0.212 &0.127 \\   
         {EA-LSS-E}~\cite{hu2023ea}  &0.776 &0.766 &0.234 &0.228 &0.278 &0.204 &0.124  \\
         {BEVFusion4D-E}~\cite{cai2023bevfusion4d}  &0.772 &0.768 &0.229 &0.229 &0.302 &0.225 &0.135 \\
         \rowcolor[RGB]{220,220,220} {MV2DFusion-E}  &\textbf{0.788}	&\textbf{0.779}	&0.237	&0.226	&0.247	&0.192	&0.119 \\
         \bottomrule

\end{tabular}

\label{tab:nus_test}
\end{center}
\end{table*}

\subsection{Implementation Details}
By default, MV2DFusion uses Faster R-CNN~\cite{FasterRCNN} with ResNet-50~\cite{ResNet} as the image-based 2D detector and FSDv2~\cite{fsdv2} as the point cloud-based 3D detector if not specified. 
We retain up to $60$ objects per image for image-based 2D detection, while allowing up to $200$ objects for point cloud-based 3D detection.
We do not modify the pipelines and hyperparameters of the modality detectors to highlight the compatibility of our method.
The fusion decoder contains 6 layers in total. 
The experiments are conducted on 8 Nvidia RTX-3090 GPUs. 
All the models are trained using AdamW~\cite{DBLP:conf/iclr/LoshchilovH19} optimizer with a weight decay of $0.01$.
Cosine annealing policy~\cite{loshchilov2016sgdr} is adopted and the initial learning rate is set to $4 \times 10^{-4}$.
The total batch size is 16 by default.

In the experiments on nuScenes \texttt{val} set, the input image resolution is set to $1600\times640$ and the voxel size is $(0.2, 0.2, 0.2)$. We use the nuImages~\cite{nuscenes2019} pre-trained weights for image-based detector and nuScenes pre-trained weights for point cloud-based detector. We freeze the point cloud-based detector to avoid overfitting. The whole model is trained for 24 epochs on nuScenes \texttt{train} set.
In the experiments on nuScenes \texttt{test} set, we replace the image-based detector by Cascade R-CNN~\cite{cascade-rcnn} with ConvNeXt-L~\cite{liu2022convnet,convnextv2} and train the whole model for 48 epochs on nuScenes \texttt{train} and \texttt{val} set for better performance.

For AV2 dataset, the input image resolution is set to $1536\times1184$ and the voxel size is $(0.2, 0.2, 0.2)$. We use the nuImages~\cite{nuscenes2019} pre-trained weights for image-based detector and AV2 pre-trained weights for point cloud-based detector. The whole model is trained for 6 epochs on AV2 \texttt{train} set.

\begin{table}[t]
\begin{center}
\caption{Performance on the nuScenes \texttt{val} sets. Our model is marked in \colorbox[rgb]{0.8627,0.8627,0.8627}{gray}. \textbf{Bold} indicates the best performance.}

\begin{tabular}{l|c|cc}
    \toprule
         {Method}  & Image Backbone  & NDS $\uparrow$ & mAP $\uparrow$  \\
         \midrule


         {TransFusion~\cite{transfusion}} &  ResNet-50  & 0.713  & 0.675
 \\
         {Autoalignv2}~\cite{autoalignv2} &CSPNet  &0.712  &0.671 \\
         {BEVFusion}~\cite{bevfusion_pku}  &Swin-T &  0.721 & 0.696 \\
         {BEVFusion}~\cite{bevfusion_mit}  &Swin-T  &0.714 &0.685  \\
         {UVTR}~\cite{li2022unifying}  &ResNet-101   &0.702 &0.654  \\
         {DeepInteraction}~\cite{yang2022deepinteraction}  &ResNet-50  &0.726 &0.699\\
         {CMT}~\cite{yan2023cross}  &V2-99 & 0.729 & 0.703\\
         {EA-LSS}~\cite{hu2023ea}  &Swin-T &0.731 &0.712 \\
         {BEVfusion4D}~\cite{cai2023bevfusion4d}  &Swin-T  &0.735  &0.720\\
         {SparseFusion}~\cite{zhou2023sparsefusion}  &Swin-T &0.728 &0.704 \\
         {FusionFormer}~\cite{hu2023fusionformer}  &V2-99  &0.741 &0.714\\
         \rowcolor[RGB]{220,220,220}{MV2DFusion}  &ResNet-50 &0.747	&0.728 \\
         \rowcolor[RGB]{220,220,220} {MV2DFusion}  &ConvNeXt-L &\textbf{0.754}	&\textbf{0.739}	 \\
         \bottomrule

\end{tabular}

\label{tab:nus_val}
\end{center}
\end{table}

\begin{table*}[t]
\begin{center}
\caption{Performance on the AV2 \texttt{val} sets. $\dagger$ is the implementation from SparseFusion \cite{sparsefusion_tusimple}. ``C'' means the input modality is camera. ``L'' means the input modality is LiDAR. ``CL'' means the input modality is camera and LiDAR. Our model is marked in \colorbox[rgb]{0.8627,0.8627,0.8627}{gray}. \textbf{Bold} indicates the best performance.}
\begin{tabular}{l|c|cc|ccc}
\toprule
Methods & Modality  &mAP $\uparrow$ & CDS $\uparrow$ & mATE $\downarrow$ & mASE $\downarrow$ & mAOE $\downarrow$ \\
\midrule
Far3D \cite{jiang2024far3d} & C   & 0.316 & 0.239 & 0.732 & 0.303 & 0.459 \\
\hline
CenterPoint \cite{centerpoint} & L  & 0.220 & 0.176 & - & - & -\\
FSD \cite{FSD} & L  &0.282 & 0.227 & 0.414 & 0.306 & 0.645\\
VoxelNeXt \cite{voxelnext} & L  & 0.307 & - & - & - & -\\
FSDv2 \cite{fsdv2} & L & 0.376 & 0.302 & 0.377 & 0.282 & 0.600\\
SAFDNet~\cite{safdnet} & L & 0.397 & - & - & - & - \\
\hline
FSF \cite{FSF} & CL  & 0.332 & 0.255 & 0.442 & 0.328 & 0.668\\
CMT \cite{yan2023cross} & CL  & 0.361 & 0.278 & 0.585 & 0.340 & 0.614 \\
BEVFusion$\dagger$ \cite{bevfusion_mit} & CL  &0.388 & 0.301 & 0.450 & 0.336 & 0.643 \\
SparseFusion \cite{sparsefusion_tusimple} & CL  & 0.398 & 0.310 & 0.449 & 0.308 & 0.677\\
\rowcolor[RGB]{220,220,220} MV2DFusion  & CL  & \textbf{0.486} & \textbf{0.395} & 0.368 & 0.267 & 0.510 \\
\bottomrule
\end{tabular}

\label{tab:av2_val}
\end{center}
\end{table*}

\subsection{Comparison with Existing Methods}
We compared MV2Dfusion with other state-of-the-art methods. The results on the nuScenes \texttt{test} set and \texttt{val} set are shown in Table~\ref{tab:nus_test} and Table~\ref{tab:nus_val}, respectively. The results on the AV2 \texttt{val} set are shown in Table~\ref{tab:av2_val}.

Table~\ref{tab:nus_test} presents the results on the nuScenes \texttt{test} set. In the single model setting, our method achieves state-of-the-art performance with 76.7\% NDS and 74.5\% mAP, outperforming all previous methods. Compared to FusionFormer~\cite{hu2023fusionformer}, we improved NDS by 1.6\% and mAP by 1.9\%. Additionally, compared to the sparse-structured SparseFusion~\cite{zhou2023sparsefusion}, we achieved improvements of 2.9\% NDS and 2.5\% mAP. Among the results with model ensembling, our model reached 78.8\% NDS and 74.5\% mAP, ranking first among all solutions. These results demonstrate the significant advantages of our model in multi-modal 3D Detection performance.

Table~\ref{tab:nus_val} presents the results on the nuScenes \texttt{val} set. Our model with a ResNet-50 backbone achieves 74.7\% NDS and 72.8\% mAP. With a ConvNeXt-L~\cite{convnextv2} backbone, it achieves 75.4\% NDS and 73.9\% mAP. Notably, our model with the ResNet-50~\cite{ResNet} backbone already outperforms the current state-of-the-art. It surpasses the FusionFormer~\cite{hu2023fusionformer} using V2-99 by 0.6\% NDS and 1.4\% mAP and outperforms SparseFusion~\cite{zhou2023sparsefusion} using SwinT~\cite{liu2021swin} by 1.9\% NDS and 2.4\% mAP. This demonstrates that our method achieves excellent performance even with a weaker backbone and achieve new state-of-the-art results with larger image backbones.

We compared the results of MV2Dfusion on the long-range AV2 dataset, shown in Table~\ref{tab:av2_val}. Our method achieves 48.6\% mAP and 39.5\% CDS, significantly surpassing previous methods. Notably, compared to the LiDAR-base state-of-the-art method, FSDv2~\cite{fsdv2}, our method achieves 10.6\% higher mAP and 9.3\% higher CDS. Compared to the multi-modal method SparseFusion~\cite{sparsefusion_tusimple}, our method outperforms by 8.8\% mAP and 8.5\% CDS, and also shows substantial improvements in mATE, mASE, and mAOE metrics over previous SOTA methods. This demonstrates the superior performance of our method on long-range scenarios.

\subsection{Ablation Study}
In this section, we perform ablation studies on the nuScenes \texttt{val} and AV2 \texttt{val} set. The experiments are conducted on the nuScenes dataset by default if not specified.

\begin{table}[t]
\setlength{\tabcolsep}{0.99mm}
\begin{center}
\caption{Ablation of flexibility on image branch.}
\begin{tabular}{c|ccc|c|cc}
\toprule
 \multirow{2}{*}{\#} & \multicolumn{3}{c|}{\textbf{Image}} & \multicolumn{1}{c|}{\textbf{LiDAR}} & \multicolumn{2}{c}{\textbf{Performance}} \\
 &resolution & backbone & detector & detector & NDS & mAP \\
\midrule
1 &704$\times$256 & ResNet-50 & YOLOX & FSDv2  & 0.740 & 0.720 \\
2 & 1600$\times$640 & ResNet-50 & YOLOX & FSDv2  & 0.745 & 0.728 \\
3 & 1600$\times$640 & ResNet-50 & Faster R-CNN & FSDv2   & 0.747 & 0.728 \\
 4 & 1600$\times$640 & ResNet-50 & Cascade R-CNN & FSDv2  & 0.747 & 0.729 \\
 5 & 1600$\times$640 & ConvNext-L & Cascade R-CNN & FSDv2  & 0.755 & 0.739 \\
\bottomrule
\end{tabular}



\label{tab:image_branch}
\end{center}
\end{table}

\subsubsection{Flexibility on Image Branch}
MV2DFusion is compatible with any type of 2D detectors and 3D detectors.
To validate the flexibility of the image branch, we conduct experiments with different detectors, image resolution and backbone, shown in Table~\ref{tab:image_branch}.
We choose 3 kinds of 2D detectors, including a single-stage detector YOLOX~\cite{yolox2021},
a two-stage detector Faster RCNN~\cite{FasterRCNN}, and a multi-stage detector Cascade RCNN~\cite{cascade-rcnn}. Rows \#2 to \#4 are the results with different 2D detectors, showing that our framework can effectively handle different 2D detectors. Additionally, comparing rows \#1 and \#2, it can be observed that a larger image resolution can bring better results. From row \#4 and row \#5, it is shown that using a more powerful image backbone can bring further improvements, demonstrating the significance of utilizing image information in the multi-modal setting.

\begin{table}[t]
\setlength{\tabcolsep}{0.99mm}
\begin{center}
\caption{Ablation of Flexibility on LiDAR branch.}
\begin{tabular}{c|ccc|c|cc}
\toprule
  \multirow{2}{*}{\#}& \multicolumn{3}{c|}{\textbf{Image}} & \multicolumn{1}{c|}{\textbf{LiDAR}} & \multicolumn{2}{c}{\textbf{Performance}} \\
 &resolution & backbone & detector & detector & NDS & mAP \\
\midrule

1&1600$\times$640 & ResNet-50 & Faster R-CNN & FSDv2   & 0.747 & 0.728 \\
2&1600$\times$640 & ResNet-50 & Faster R-CNN & VoxelNeXt  & 0.735 & 0.727 \\
3&1600$\times$640 & ResNet-50 & Faster R-CNN & Transfusion-L  & 0.748 & 0.728 \\
\bottomrule
\end{tabular}
\label{tab:LiDAR_branch}
\end{center}
\end{table}

\subsubsection{Flexibility on LiDAR branch}
In Table~\ref{tab:LiDAR_branch} we compare different detectors on the LiDAR branch. We experiment with mainstream point cloud-based 3D detectors, including two sparse detectors FSDv2~\cite{fsdv2} and VoxelNext~\cite{voxelnext}, and one dense BEV detector TransFusion-L~\cite{transfusion}. The experimental results show that various LiDAR detectors can be well adapted to our framework. Among them, FSDv2~\cite{fsdv2} and TransFusion-L~\cite{transfusion} achieve slightly better results. 
To embrace the sparse nature of the proposed fusion strategy, we employ FSDv2 as the default point cloud-based detector.

\begin{table}[t]
\begin{center}
\caption{Ablation of modality robustness. ``C'' means the input modality is camera. ``L'' means the input modality is LiDAR. ``CL'' means the input modality is camera and LiDAR.}
\begin{tabular}{c|cc|cc}
\toprule
\multirow{2}{*}{\#}& \multicolumn{2}{c|}{\textbf{Input Modality}}  & \multicolumn{2}{c}{\textbf{Performance}} \\
&train & val & NDS & mAP \\
\midrule
1&LC & LC & 0.747 & 0.728 \\
\hline
2&LC & C & 0.469 & 0.400 \\
3&LC & L & 0.687 & 0.615 \\
4&C & C & 0.547 & 0.460 \\
5&L & L & 0.709 & 0.665 \\
\hline
6&LC/L/C & C & 0.523 & 0.441 \\
7&LC/L/C & L & 0.713 & 0.663 \\
8&LC/L/C & LC & 0.748 & 0.731 \\
\bottomrule
\end{tabular}

\label{tab:modality_robustness}
\end{center}
\end{table}

\begin{table}[t]
\begin{center}
\caption{Ablation of modality queries. ``distribution'' means the query is formulated by a probability distribution, while ``point'' means the query is formulated by a 3D center point.}
\begin{tabular}{c|c|cc|cc}
\toprule
\multirow{2}{*}{\#} & \multirow{2}{*}{\textbf{Dataset}} & \multicolumn{2}{c|}{\textbf{Query Type}} & \multicolumn{2}{c}{\textbf{Performance}} \\
& & image & LiDAR & NDS/CDS & mAP \\
\midrule
1& \multirow{4}{*}{nuScenes}  & point & point& 0.743 & 0.725 \\
2& & - & point & 0.737 & 0.716 \\
3& & point & - & 0.733 & 0.707 \\
4&& distribution& point & 0.747 & 0.728 \\
\hline
5&\multirow{2}{*}{AV2} & point & point & 0.373 & 0.464 \\
6& & distribution & point & 0.395 & 0.486 \\
\bottomrule
\end{tabular}
\label{tab:query_ablation}
\end{center}
\end{table}

\subsubsection{Ablation of Modality Robustness}
Modality robustness is an important aspect of fusion methods to maintain functionality when modality sensors fail.
In Table~\ref{tab:modality_robustness}, we validate the modality robustness of our model. Row \#1 lists the results when both LiDAR and camera are used for training and evaluation, which is our default setting. 
Rows \#2 and \#3 present the results when one modality is missing during evaluation, despite the model being trained with both modalities. In contrast, rows \#4 and \#5 show the results when the same modality is used for both training and evaluation. The performances in rows \#2 and \#3 exhibit noticeable degradations compared to rows \#4 and \#5, suggesting substantial risks in the event of sensor failure.
To alleviate this problem, we can use a mixed combination of modalities during training, i.e., the input modality is randomly selected from $[\emph{camera}, \emph{LiDAR}, \emph{camera and LiDAR}]$ with probabilities $[0.2, 0.1, 0.7]$. As shown in rows \#6 and \#7, when LiDAR or camera sensors are missing in this case, the model still produces decent performance. These results indicate that the modality robustness can be effectively improved with more single-modality training samples.
Comparing row \#8 and row \#1, mixed modality training also slightly boosts the performance in the multi-modal scenario.

\subsubsection{Ablation of Modality Queries}
We conduct an ablation study on modality queries, and Table~\ref{tab:query_ablation} presents the results.
First, we verify the impact of the query formulation on the nuScenes and AV2 datasets. Note that the point formulation of image queries remains consistent with MV2D~\cite{mv2d}, where each image query is represented by an estimated 3D center point. 
In the nuScenes dataset, as shown in rows \#1 and \#4, the distribution formulation outperforms the point formulation by 0.4\% NDS and 0.3\% mAP.
In the AV2 dataset, the advantage is enlarged to 2.2\% CDS and 2.4\% mAP, according to rows \#5 and \#6. This remarkable performance gain in the AV2 dataset is attributed to the long-range perception setting where the depth estimation from images is more challenging. In this case, the point formulation of image queries can introduce large inaccuracies. This comparison validates the necessity of appropriately treating different modalities.

We also evaluate the performance when input with only point cloud and image queries respectively in rows \#1 to \#3. Compared to using both modality queries, it is evident that using queries from a single modality results in performance degradation. 
These results prove that the object semantics from each modality do have a substantial effect on the overall performance.

\begin{table*}[t]
\setlength{\tabcolsep}{0.985mm}
\begin{center}
\caption{Comparison of memory cost on different datasets.}

\begin{tabular}{c|ccccc|ccccc}
\toprule
\multirow{2}{*}{\textbf{Method}} & \multicolumn{5}{c|}{\textbf{nuScenes}} & \multicolumn{5}{c}{\textbf{AV2}} \\
 & point range & pillar scale & BEV scale & resolution & memory & point range & pillar scale & BEV scale & resolution & memory \\
\midrule
BEVFusion & \makecell[l]{\\x: [-54.0, 54.0] \\y: [-54.0, 54.0] \\z: [-5.0, 3.0]} & - & [180, 180] & 1600$\times$640 & 22099MiB & \makecell[l]{\\x: [-204.0, 204.0] \\y: [-204.0, 204.0] \\z: [-3.0, 3.0]} & - & [680, 680] & 1536$\times$1184 & OOM ($>$24.5G) \\
\hline
\makecell{\\BEVFusion*\\(BEVPool V2)} & \makecell[l]{\\x: [-54.0, 54.0] \\y: [-54.0, 54.0] \\z: [-5.0, 3.0]} & - & [180, 180] & 1600$\times$640 & 11637MiB & \makecell[l]{\\x: [-204.0, 204.0] \\y: [-204.0, 204.0] \\z: [-3.0, 3.0]} & - & [680, 680] & 1536$\times$1184 & 23465MiB \\
\hline
MV2Dfusion & \makecell[l]{\\x: [-54.4, 54.4] \\y: [-54.4, 54.4] \\z: [-5.0, 3.0]} & 8814 & - & 1600$\times$640 & 6095MiB & \makecell[l]{\\x: [-204.8, 204.8] \\y: [-204.8, 204.8] \\z: [-3.2, 3.2]} & 8750 & - & 1536$\times$1184 & 8215MiB \\
\bottomrule
\end{tabular}

\label{tab:memory_cost}
\end{center}
\end{table*}

\begin{table*}[t]
\setlength{\tabcolsep}{1.7mm}
\begin{center}
\caption{Comparison of inference speed on different datasets.}
\begin{tabular}{c|ccc|ccc}
\toprule
\multirow{2}{*}{\textbf{Method}} & \multicolumn{3}{c|}{\textbf{nuScenes}} & \multicolumn{3}{c}{\textbf{AV2}} \\
 & point range &  resolution & FPS & point range &  resolution & FPS \\
\midrule
\makecell{BEVFusion* \\ (BEVPool V2)} & \makecell[l]{\\x: [-54.0, 54.0] \\y: [-54.0, 54.0] \\z: [-5.0, 3.0]} & 704$\times$256 & 4.4 & \makecell[l]{\\x: [-204.0,204.0] \\y: [-204.0,204.0] \\z: [-3.0, 3.0]} & 1536$\times$1184 & 0.9 \\
\hline
MV2Dfusion & \makecell[l]{\\x: [-54.4, 54.4] \\y: [-54.4, 54.4] \\z: [-5.0, 3.0]} & 704$\times$256 & 5.5 & \makecell[l]{\\x: [-204.8,204.8] \\y: [-204.8,204.8] \\z: [-3.2, 3.2]} & 1536$\times$1184 & 2.0 \\
\bottomrule
\end{tabular}
\label{tab:inference_speed}
\end{center}
\end{table*}

\begin{table}[t]
\begin{center}
\caption{Ablation of decoder structure. }
\begin{tabular}{c|c|cc|cc}
\toprule
\multirow{2}{*}{\#}&\multirow{2}{*}{\textbf{Layer Num}} & \multicolumn{2}{c|}{\textbf{Structure}} & \multicolumn{2}{c}{\textbf{Performance}} \\
& & self-attn & cross-attn & NDS & mAP \\
\midrule
1&1 & \checkmark & \checkmark & 0.742 & 0.726 \\
2&2 & \checkmark & \checkmark & 0.746 & 0.728 \\
3&3 & \checkmark & \checkmark & 0.746 & 0.729 \\
4&6 & \checkmark & \checkmark & 0.747 & 0.728 \\
5&6 & \checkmark &            & 0.740 & 0.715 \\
\bottomrule
\end{tabular}
\label{tab:decoder_structure}
\end{center}
\end{table}

\subsubsection{Decoder Structure}
We conduct experiments on different structures of the decoder, with the results presented in Table~\ref{tab:decoder_structure}. In rows \#1 to \#4, we examine the impact of varying the number of layers on the final results. It can be observed that, overall, the accuracy increases with more layers. However, when comparing the results of using 3 layers versus 6 layers, the performance improvement is relatively minimal, suggesting that additional layers contribute less to performance gains beyond 3 layers.

Additionally, we evaluate the influence of the cross-attention layers within the decoder. As given by rows \#4 and \#5, our findings indicate that removing the cross-attention layer results in a 0.7\% decrease in NDS and a 1.3\% decrease in mAP, highlighting the efficacy of the cross-attention layers in enhancing the decoder's performance. On the other hand, despite the removal of the computationally intensive cross-attention layers, the model still maintains 74.0\% NDS and 71.5\% mAP. This outcome indicates that our framework can achieve decent performance with query-level fusion alone, thereby proving the effectiveness and robustness of the multi-modal query design.

\begin{table}[t]
\begin{center}
\caption{Ablation of history information.}
\begin{tabular}{c|cc|cc}
\toprule
\multirow{2}{*}{\#}&\multicolumn{2}{c|}{\textbf{History}}
 & \multicolumn{2}{c}{\textbf{Performance}} \\
&frame num & query num & NDS & mAP \\
\midrule

1&- & - & 0.729 & 0.710 \\
2&6 & 128 & 0.745 & 0.727 \\
3&6 & 256 & 0.747 & 0.728 \\
4&12 & 256 & 0.749 & 0.731 \\
\bottomrule
\end{tabular}
\label{tab:temporal_information}
\end{center}
\end{table}

\subsubsection{History Information}
In Table~\ref{tab:temporal_information}, we illustrate the impact of history information on the results. Row \#1 presents the results without incorporating history information. Row \#2 incorporates information from the previous 6 frames, resulting in a significant improvement of 1.6\% NDS and 1.7\% mAP. 
The comparison between row 2 and row 3 reveals that augmenting the number of historical queries also enhances performance. Additionally, extending to 12 history frames yields further improvements of 0.2\% NDS and 0.3\% mAP, indicating that the model can benefit from more history information.

\begin{figure*}[t!]
	\centering
	\vspace{-2mm}
	\includegraphics[width=1.0\textwidth]{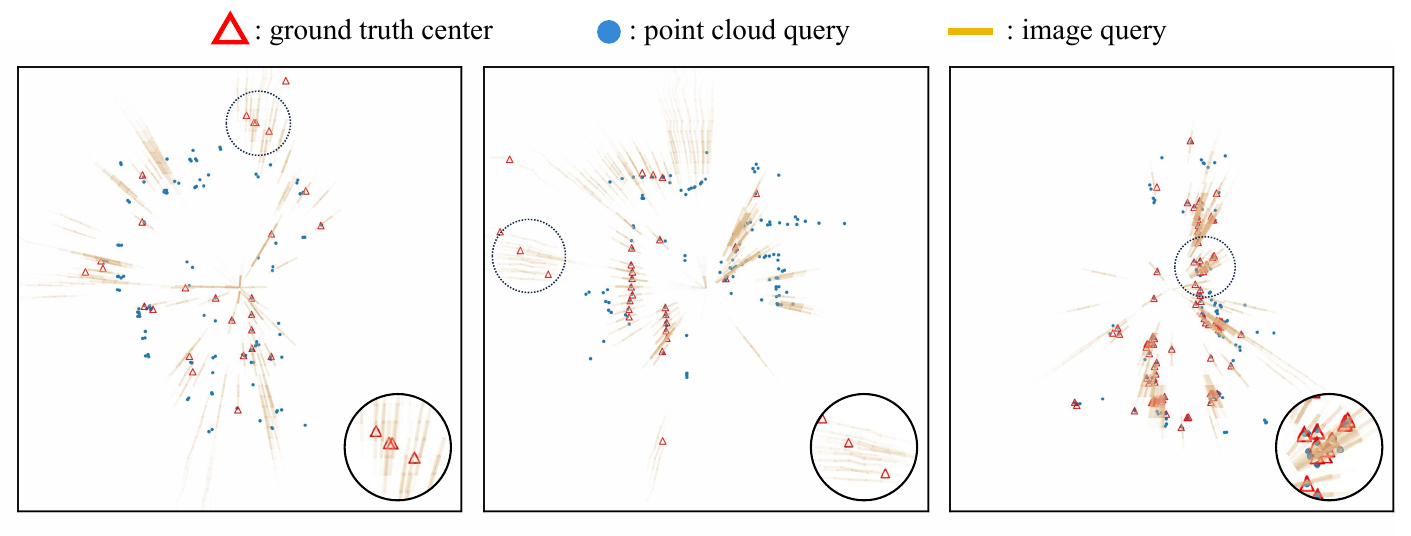}
	\caption{Visualization of modality queries. The ground truth objects are represented by \textcolor{black}{red triangles} situated at their center points. The point cloud queries are represented by \textcolor{black}{blue circles} situated at the center points $\mathbf{r}_{pc}$.  The image queries are represented by \textcolor{black}{orange line segments}, where the placement of the line segment signifies the sampling positions $\mathbf{s}^{img}$, the intensity of the color saturation indicates the probability distribution $\mathbf{u}^{img}$.}
	\vspace{-4mm}
	\label{fig:vis_query}
\end{figure*}

\subsubsection{Memory Cost}
We compare the memory cost with a typical feature-level fusion method, BEVFusion, on nuScenes and AV2. The results are shown in Table~\ref{tab:memory_cost}.
To demonstrate the efficiency of our method, we additionally implement BEVFusion with an advanced technique BEVPoolv2\cite{bevpoolv2} which greatly accelerates the view transform process for comparison. 
The collected statistics are averaged on the validation set.

Our model, which produced only sparse pillar features in point cloud branch, has smaller scales of features compared to the dense BEV-based method BEVFusion (8814 pillars compared to 180*180 BEV grids). We can see that our model has significantly lower memory usage. Specifically, compared to BEVFusion with BEVPool V2, our model requires only about half the memory.
On the long-range AV2 dataset, our pillar scale is much smaller than BEVFusion's dense BEV scale. By maintaining the same resolution, we compare the memory cost of the models. BEVFusion exceeds 24.5GB, while BEVFusion with BEVPool V2 requires 23.4GB of memory. Our model, on the other hand, needs only 8.2GB of memory, which is approximately 35\% of the memory cost of BEVFusion with BEVPool V2.
From the results, our method shows a significant advantage in memory cost compared to feature-level fusion methods, demonstrating the superiority of our sparse fusion strategy.

\begin{figure}[t!]
	\centering
	\includegraphics[width=0.48\textwidth]{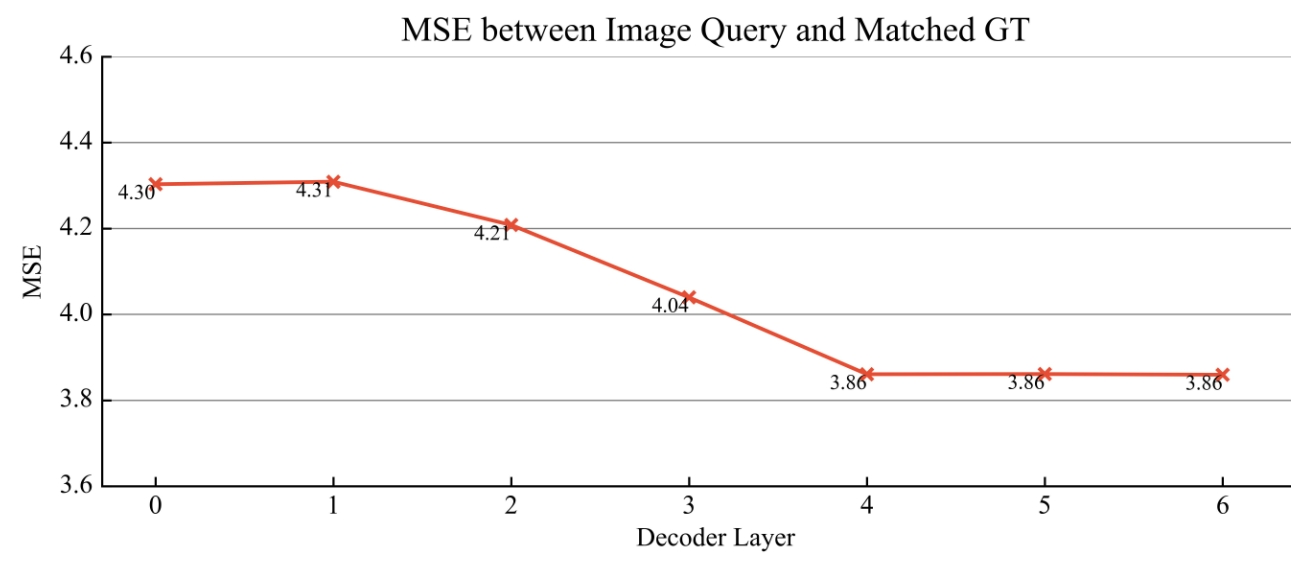}
	\caption{The MSE is calculated between image queries and matched GTs. Specifically, we exclude all the point cloud queries and perform one-to-one matching~\cite{kuhn1955hungarian} between image queries and GTs for calculation.}
	\label{fig:vis_mse}
	\vspace{-4mm}
\end{figure}

\subsubsection{Inference Speed}
In Table~\ref{tab:inference_speed}, we test the inference speed under two extreme settings: small image resolution with a short perception range (nuScenes), and large image resolution with a long perception range (AV2). The FPS is evaluated with a batch size of 1.
The collected statistics are averaged on the validation set.

On the nuScenes dataset, while having nearly the same point range and image resolution, our method is $25\%$ faster than BEVFusion with BEVPool V2 (5.5 FPS vs. 4.4 FPS). On the long-range AV2 dataset, our sparse model exhibits a more significant speed advantage. In this more resource-intensive scenario, the FPS of our method is twice that of BEVFusion with BEVPool V2 (2.0 FPS vs. 0.9 FPS).

\begin{figure*}[!tbp]
	\centering
	\includegraphics[width=1.0\textwidth]{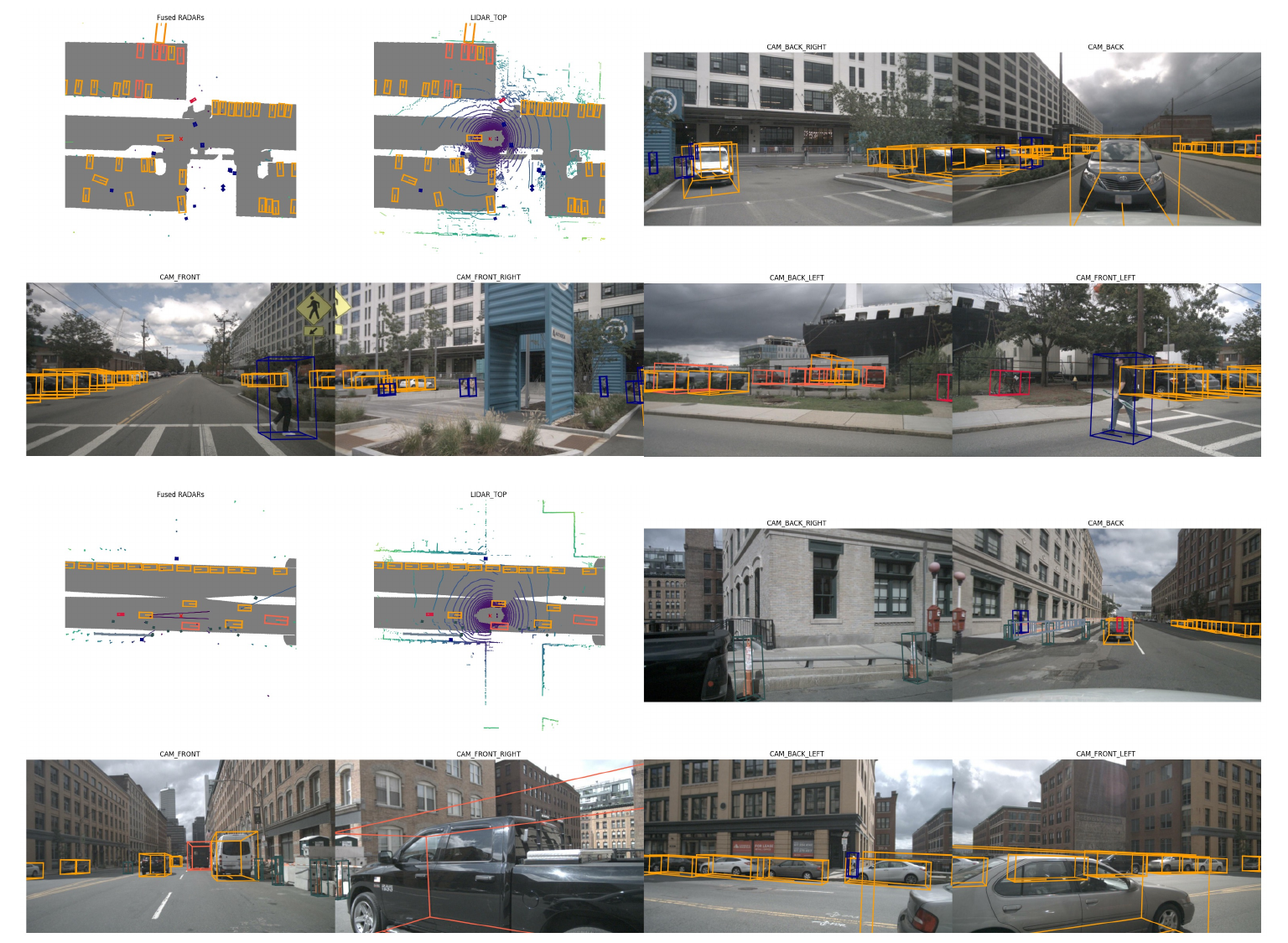}
	\caption{Visualization of prediction results. The results are drawn from nuScenes validation set.}
	\vspace{-2mm}
	\label{fig:vis_result}
\end{figure*}

\subsection{Qualitative Results}
We visualize the queries in Figure~\ref{fig:vis_query} to deliver a lucid understanding of the complementary nature of modality-specific object semantics. The ground truth objects are represented by \textcolor{black}{red triangles} situated at their center points. The point cloud queries are represented by \textcolor{black}{blue circles} at the center points $\mathbf{r}_{pc}$.  The image queries are represented by \textcolor{black}{orange line segments}, where the placement of the line segment signifies the sampling positions $\mathbf{s}^{img}$. The saturation of the color indicates the probability distribution $\mathbf{u}^{img}$.
From the visualization, despite the sparsity in 3D space, modality queries are inclined to be distributed around objects, allowing them to still accurately identify the objects.
Furthermore, these two kinds of queries exhibit their different capabilities to help localize objects.
Image queries are able to recall some objects missed by point cloud queries due to excessively incomplete point clouds (e.g., farther distances). Meanwhile, the 3D information contained in point cloud queries assists in accurately locating and distinguishing objects even in crowded scenes when the image queries struggle to do so.

To validate our query calibration design, we compute the mean squared error (MSE) between the positions of the image queries and the matched ground-truths, as shown in Figure~\ref{fig:vis_mse}. We observe that the MSE decreases at deeper decoder layers, which indicates that our query calibration effectively refines the position of the image query progressively.

We also visualize the prediction results on nuScenes validation set in Figure~\ref{fig:vis_result}. It can be seen that MV2DFusion can detect various objects including persons, vehicles, barriers, traffic cones, and so on.

\section{Conclusion}
In this paper, we present MV2DFusion, a general and efficient multi-modal framework for object detection. Continuing the object as query approach of its predecessor MV2D, MV2DFusion introduces modality-specific object detectors to generate object queries. Then a decoder with cross attention is adopted to aggregate features from both LiDAR and image space and predict final results. 
This design allows for compatibility with various image and LiDAR detectors and enables long-range detection due to the sparsity inherent in the fusion process. We validate the framework across diverse scenarios, demonstrating superior performance compared to recent state-of-the-art methods on both the nuScenes and AV2 datasets. We hope this design will advance the development of multi-modal object detection frameworks in both research and practical applications.


%





\ifCLASSOPTIONcaptionsoff
  \newpage
\fi



\bibliographystyle{IEEEtran}
\bibliography{trans}

\end{document}